\documentclass[letterpaper, 10 pt, conference]{ieeeconf}  

\IEEEoverridecommandlockouts                              

\usepackage{authblk}
\usepackage{amsmath}
\usepackage{amssymb}
\usepackage{subcaption}
\usepackage{float}  
\usepackage{colortbl} 
\usepackage{booktabs}
\usepackage{soul}
\usepackage{tcolorbox}
\usepackage{kotex} 
\usepackage{multicol} 
\usepackage{wrapfig}
\usepackage{makecell} 
\usepackage{tabularx} 
\usepackage{xcolor}
\usepackage{cite}
\usepackage[colorlinks=true,linkcolor=blue,citecolor=green,urlcolor=magenta]{hyperref}

\definecolor{deeppink}{RGB}{255,20,147}
\definecolor{deepblue}{RGB}{22,83,126}
\definecolor{lightblue}{RGB}{173,216,230}
\definecolor{lightviolet}{RGB}{238,130,238}
\definecolor{lightred}{RGB}{255,182,193}
\definecolor{lightgreen}{RGB}{144,238,144}
\definecolor{lightyellow}{RGB}{255,255,224}
\definecolor{fluorescentpink}{RGB}{240,168,158} 
\definecolor{fluorescentgreen}{RGB}{160,219,142} 
\definecolor{fluorescentpurple}{RGB}{204,153,255} 
\definecolor{fluorescentorange}{RGB}{255,165,0} 
\definecolor{fluorescentyellow}{RGB}{239,244,131}

\title{\LARGE \bf
MGHanD: Multi-modal Guidance for authentic Hand Diffusion
}

\author{Taehyeon Eum$^{1,2}$, Jieun Choi$^{1,2}$, Tae-Kyun Kim$^{1}$
\thanks{$^{1}$Authors are with the KCVL Lab, KAIST, Daejeon, Republic of Korea. Emails:\{taehyeon\_eum, jichoi0101, kimtaekyun\}@kaist.ac.kr}
\thanks{$^{2}$Taehyeon Eum and Jieun Choi were supported by KT Corporation through an academic-industry cooperation program.}
}

\begin{document}

\maketitle

\begin{figure*}[htbp]
    \centering
    \includegraphics[width=\textwidth]{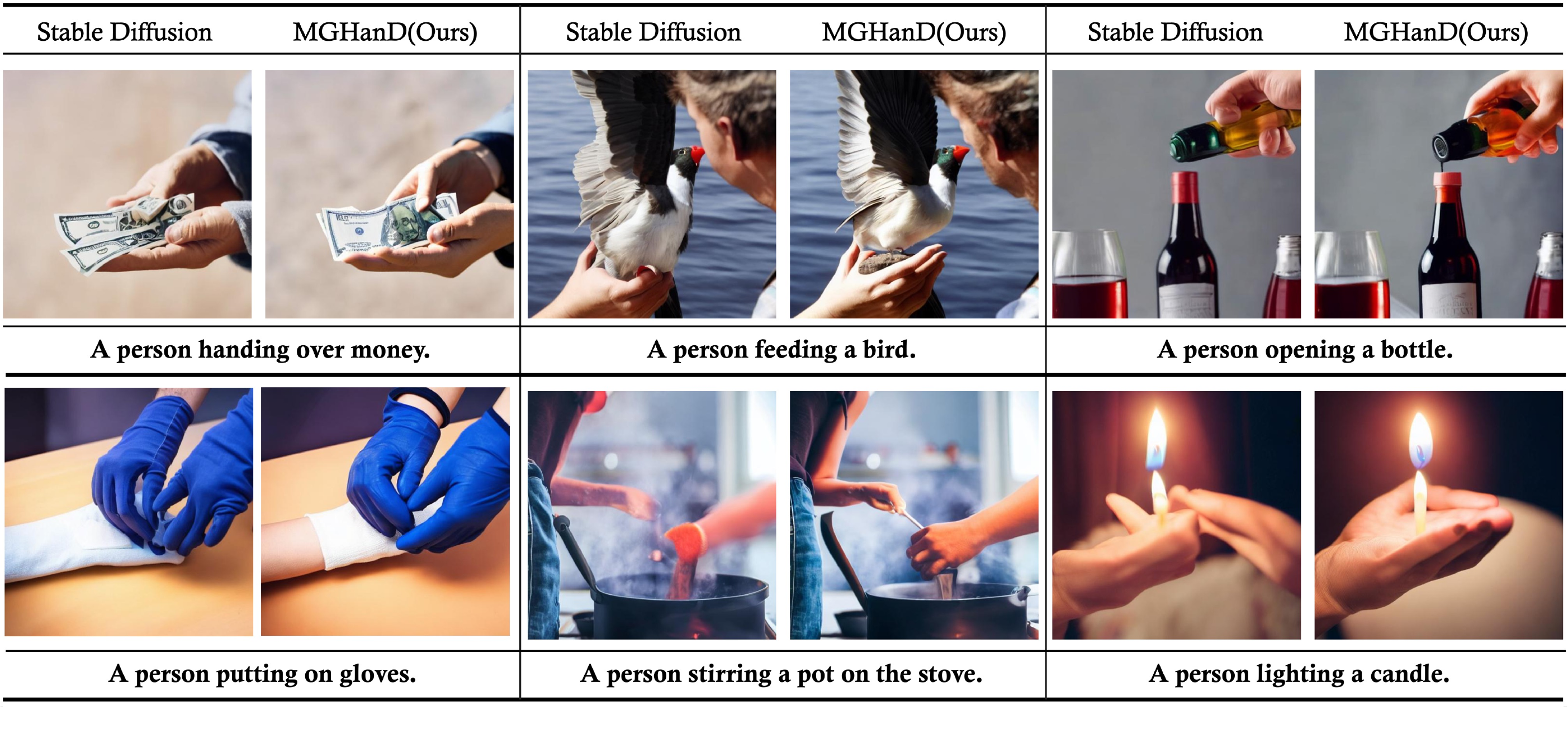} 
    \vspace{-27pt} 
    \caption{\textbf{Comparison of Stable Diffusion (left) and our method, MGHanD (right)}, for text-to-image diffusion in hand-object interaction scenarios. While Stable Diffusion often produces malformed hands with anatomical inconsistencies or blurred details, MGHanD approach refines hand articulation and pose accuracy without compromising Stable Diffusion's overall visual style. Our method preserves the image composition and aesthetic while significantly improving the realism of hand-object interactions.}
    \label{fig:main_image}
\end{figure*}
\vspace{-15pt}


\begin{abstract}

Diffusion-based methods have achieved significant successes in T2I generation, providing realistic images from text prompts. Despite their capabilities, these models face persistent challenges in generating realistic human hands, often producing images with incorrect finger counts and structurally deformed hands. MGHanD addresses this challenge by applying multi-modal guidance during the inference process. For visual guidance, we employ a discriminator trained on a dataset comprising paired real and generated images with captions, derived from various hand-in-the-wild datasets. We also employ textual guidance with LoRA adapter, which learns the direction from `hands' towards more detailed prompts such as `natural hands', and `anatomically correct fingers' at the latent level. A cumulative hand mask which is gradually enlarged in the assigned time step is applied to the added guidance, allowing the hand to be refined while maintaining the rich generative capabilities of the pre-trained model. In the experiments, our method achieves superior hand generation qualities, without any specific conditions or priors. We carry out both quantitative and qualitative evaluations, along with user studies, to showcase the benefits of our approach in producing high-quality hand images.

\end{abstract}

\section{INTRODUCTION}

Text-to-image (T2I) diffusion models, including well-known Stable Diffusion\cite{rombach2022high} SDXL\cite{podell2023sdxl}, DALL-E\cite{ramesh2021zero}, and Imagen\cite{saharia2022photorealistic}, have made significant advancements in the generation of high-quality, realistic images from text prompts. These models demonstrate a strong ability to generate visually coherent and detailed outputs that align closely with textual descriptions. Despite recent advances, generating realistic human hands remains a significant challenge for T2I diffusion models. Human hands are highly complex structures with intricate anatomical details, including 16 joints and 27 degrees of freedom, making it difficult to generate realistic images. Additionally, it fails to capture the complex interactions between hand structure and function, resulting in unnatural hand-object interactions and unrealistic poses. As illustrated in Fig.~\ref{fig:main_image}, generative models frequently produce images where hands are malformed, with issues such as incorrect numbers of fingers or distorted shapes. These persistent issues highlight the need for advancements to improve the realism of hand representations in text-to-image generative models, addressing both the structural and contextual complexities involved.

The accurate generation of human hands is not only vital for visual realism but also holds significant implications for robotics. In robotic manipulation and human-robot interaction, understanding and replicating human hand configurations are essential for developing dexterous robotic hands and improving robots’ ability to interpret human gestures. For instance, advancements in T2I models that can generate precise hand images can aid in training robotic systems to recognize and mimic complex hand movements, thereby enhancing their manipulation capabilities and interaction proficiency. Moreover, integrating such models into robotic learning frameworks can facilitate the development of more adaptable and efficient robots capable of performing intricate tasks in unstructured environments.

Recent works have attempted to address the challenges of hand generation by developing specialized techniques tailored for this task. For example, \cite{lu2023handrefiner, pelykh2024giving} suggests using 2D hand conditions such as meshes, skeletons and bounding-boxes for enhancing throughout the diffusion process. Despite these advancements, current approaches fail to ensure consistent hand detection across diverse scenarios, particularly when hands appear deformed. In such cases, hand detection models \cite{lugaresi2019mediapipe, lin2021mesh, sun2022onepose} often struggle to accurately identify and segment the hands. This limitation leads to a critical issue where the model fails to refine the hand images altogether.

To address these persistent issues, we propose a novel framework called MGHanD, which applies multi-modal guidance focused on the hand region to generate realistic hand images from text prompts, without requiring additional fine-tuning of pre-trained weights. Our method incorporates three key components: \textit{Visual Guidance}, \textit{Textual Guidance}, and \textit{a Cumulative Hand Mask} to enhance hand generation.

The \textit{\textbf{Visual Guidance}} utilizes discriminator\cite{sauer2023stylegan} trained on two real-world hand image datasets\cite{kapitanov2024hagrid,Shan20}. Existing datasets often lack detailed hand-object interaction labels. To resolve this challenge, we employ a Visual Language Model (VLM)\cite{llava2023} to generate image captions that focus on hand-centric, motion-based descriptions, such as \textit{"A person opening a jar of cookies in a kitchen"}. We then create a synthetic image-caption-paired dataset, enabling the discriminator to effectively output realistic hand features and refine hand image generation through the latent diffusion sampling process.

The \textit{\textbf{Textual Guidance}} is inspired by the ConceptSlider\cite{gandikota2023concept} approach, utilizing a LoRA adapter\cite{hu2021lora} to fine-tune the model. LoRA adaptation steers the model’s generation path from generic terms such as  \textit{`hands'} to refined descriptors such as \textit{`natural hands'} or \textit{`five fingers hands'}, effectively enhancing the realism of the hand images. 

Additionally, our framework utilizes a \textit{\textbf{Cumulative Hand Mask}} generated by hand detection model Mediapipe \cite{lugaresi2019mediapipe} at each step of the diffusion process. The Cumulative Hand Mask ensures the precise application of guidance to hand regions while maintaining the overall image style.

Consequently, by applying multi-modal guidance within the cumulative hand mask region, our approach effectively refines the hand generation process during diffusion, focusing on adjusting only the hand while preserving key visual elements such as morphology, color, and background in the image, as shown in Fig.~\ref{fig:main_image}. Through comprehensive qualitative and quantitative studies, MGHanD has been shown to outperform existing methods\cite{lu2023handrefiner, rombach2022high, gandikota2023concept}, demonstrating significant improvements in hand image generation aligned with textual prompts.

\section{Related Work}
\textbf{Image Synthesis Diffusion Models.} Text-to-image generation has rapidly evolved with the introduction of deep generative models, culminating in state-of-the-art performance achieved by diffusion-based approaches \cite{rombach2022high,balaji2022ediff, saharia2022photorealistic, ramesh2022hierarchical}. Specifically, Latent Diffusion Models (LDM) \cite{rombach2022high} perform the diffusion process within a latent image space, which greatly reduces computational demands. In this work, we use Stable Diffusion \cite{rombach2022high}, a well-known foundation model of LDM, which uses a pre-trained language encoder such as CLIP \cite{radford2021learning} to encode the text prompt into a lower-dimensional latent space. In addition to text-only input, many models have been developed to handle more diverse modalities as inputs, including Textual-Inversion, DreamBooth, and Custom-Diffusion\cite{gal2022image , ruiz2023dreambooth, kumari2023multi}, which allow for customizing diffusion models using images. Additionally, models like GLIGEN, Composer, T2I-Adaptor, and ControlNet\cite{li2023gligen, huang2023composer, mou2024t2i, zhang2023adding } incorporate modalities that provide positional information. Diffusion models are also widely used in image editing, with works such as SDEdit, iEdit, and InstructPix2Pix \cite{meng2021sdedit, bodur2024iedit, bodur2024prompt, brooks2023instructpix2pix} being prominent examples. Our model also follows a similar approach in that it retains as much of Stable Diffusion’s image generation capability as possible while focusing on the target area, which in our case is the hand.

\textbf{Diffusion Guidance Method.} The Diffusion Guidance Method is a relatively recent approach for controlling generation in diffusion models without altering the pre-trained weights. One of the most well-known methods in this category is Classifier-Free Guidance\cite{ho2022classifier}, which significantly improves generation quality by utilizing unconditional guidance, without requiring additional model training. Additionally, studies such as \cite{hertz2022prompt, tumanyan2023plug} use cross-attention guidance between target words and images to modify target elements of images. Furthermore, the Universal Guidance for Diffusion(UGD) \cite{bansal2023universal} demonstrated the broader applicability of using guidance from multiple models to direct and control generation, further proving the practical utility of diffusion guidance. For these reasons, we apply the guidance of a Hand Discriminator to our model following the approach of Universal Guidance for Diffusion.

\begin{figure*}[ht]
    \centering
    \includegraphics[width=0.98\textwidth,height=0.4\linewidth]{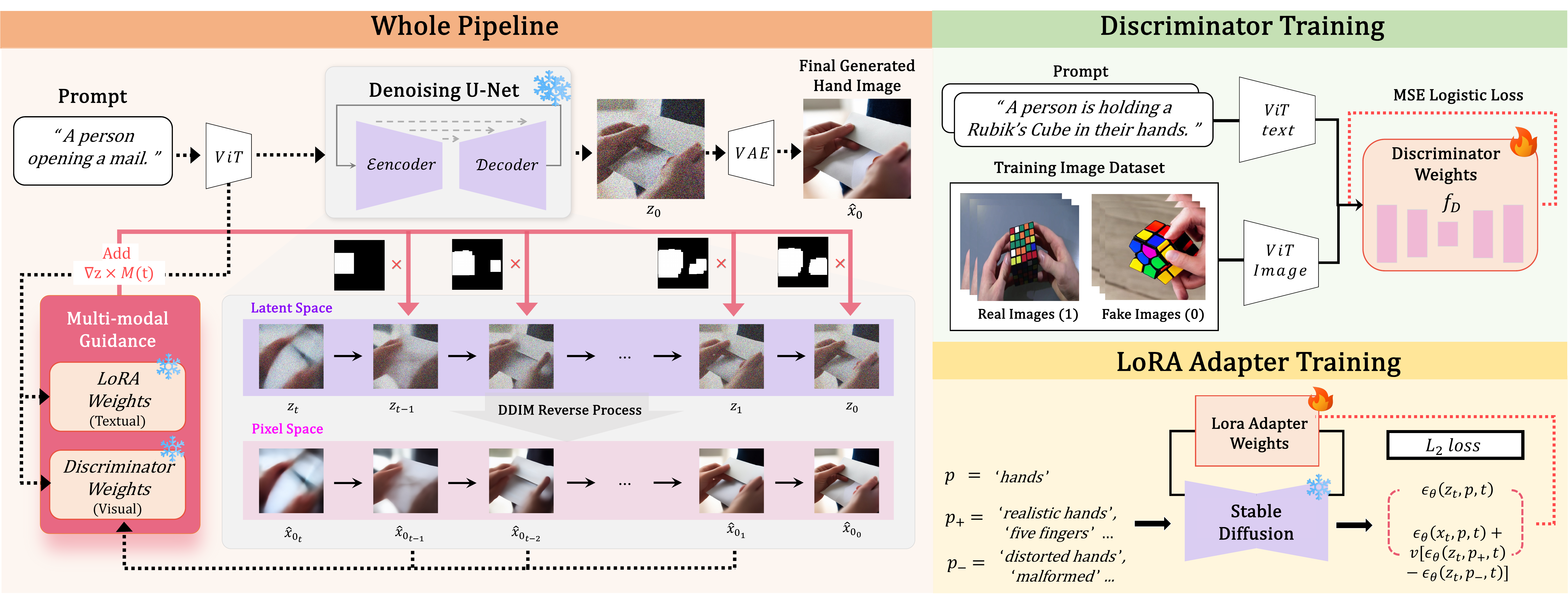} 
    \caption{\textbf{Overview of proposed MGHanD framework.} Left side: Whole Pipeline demonstrates the end-to-end process, featuring a Denoising U-Net $\epsilon_\theta$ with multi-modal guidance. The pipeline applies visual (Discriminator) and textual (LoRA) guidance weights to model with cumulative hand region masks, allowing precise control over hand features while maintaining overall image coherence. Top right: Discriminator Training module employs MSE logistic loss on real and fake hand images to enhance realism. Bottom right: LoRA Adapter Training utilizes a stable diffusion backbone for efficient fine-tuning to hand-specific prompts, employing L2 loss to optimize the adapter weights and ensure accurate text-to-hand alignment.}
    \vspace{-3mm}
    \label{fig:whole_pipeline}
\end{figure*}

\textbf{Hand-Specialized Diffusion Models.} Despite significant advancements, generating accurate hand images remains one of the most challenging tasks in image synthesis due to the complex structure of the hand and wide range of poses. Recent research has introduced various methods to address this problem\cite{zhang2023adding, pelykh2024giving, lu2023handrefiner, gandikota2023concept, narasimhaswamy2024handiffuser, wang2024rhands}. HanDiffuser\cite{narasimhaswamy2024handiffuser} employs two diffusion models: one to predict MANO\cite{romero2022embodied} parameters, an off-the-shelf hand pose model, and another to generate images using these parameters and input prompts. This dual-stage diffusion process increases computational load. ConceptSlider\cite{gandikota2023concept}, on the other hand, uses a set of prompts and trains a LoRA adapter to fine-tune hand generation. However, this can occasionally lead to imprecise directionality, resulting in slightly less detailed images. Alternatively, models like ControlNet, HumanSD, and T2I-Adapter\cite{zhang2023adding, ju2023humansd, mou2024t2i} use skeleton conditioning to naturally express human arms, legs, and hands. For example, HandRefiner\cite{lu2023handrefiner} uses a depth map of the hand to impaint, while \cite{pelykh2024giving} employs skeleton conditioning to separately generate hands and other body parts with ControlNet’s pre-trained weights. Although these models demonstrate excellent hand generation capabilities, generating hands separately from the other body parts sometimes results in discontinuities at the junctions and style differences between the hand and background. 


\vspace{1.5mm}
 Our method introduces a novel end-to-end framework that combines a LoRA adapter for Textual Guidance and a Hand Discriminator for Visual Guidance, using a cumulative mask approach to capture precise hand areas. By merging guidance with pre-trained weights at inference, we maintain the original style without adding extra training or inference overhead, ensuring both efficiency and accuracy.

\section{Methodology}
\label{sec:Methodology}

 A general overview of the proposed framework is illustrated in Fig.~\ref{fig:whole_pipeline} Given a text input, \textbf{MGHanD}  $\epsilon_{\theta_{M}}$ generates well-formed hand images. Our method incorporates two complementary forms of guidance within Stable Diffusion\cite{rombach2022high}. The first is guidance $\epsilon_{\theta_{V}}$ through a discriminator $f_D$, which takes as input $x_{0_t}$ from pixel space and provides robust hand correction visually. The second is guidance $\epsilon_{\theta_{T}}$ via a LoRA adaptor, trained to refine hand generation using a variety of text prompts $p_{+_i}$ and $p_{-_j}$. During the generation time steps, a hand mask $M_t$ is applied to maintain the fidelity of Stable Diffusion’s generative capabilities while ensuring that hand features are suitably enhanced. The following sections present a more detailed explanation of the inductive biases in \textbf{MGHanD}, covering Visual Discriminator Guidance, Textual LoRA Adapter Guidance, and the Cumulative Hand Mask technique, in order.
\vspace{1.mm}


\subsection{Preliminaries}
\textbf{Diffusion models} consist of a T-step forward process, which gradually adds Gaussian noise to a clean data point \(z_0\) in latent space, and a T-step reverse process that aims to denoise the noisy data. Mathematically, after \(t\) steps, a noisy data point \(z_t\) is produced from \(z_0\). The diffusion model \(\epsilon_\theta\) learns to predict the noise added in the forward process, and it is trained to approximate \(\epsilon\) given \(z_t\) and time step \(t\). 
\begin{equation}
\epsilon_\theta(z_t, t) \approx \epsilon = \frac{z_t - \sqrt{\alpha_t} z_0}{\sqrt{1 - \alpha_t}}.
\end{equation}
The reverse process is typically modeled as a Gaussian distribution, which attempts to recover \(z_0\) from \(z_t\). DDIM\cite{song2020denoising}, modify the reverse process by estimating a clean data point \(\hat{z}_0\) in the latent space and then using it for sampling the next step. It samples $z_{t-1}$ from $q(z_{t-1}|z_{t}, \hat{z}_0)$ by replacing unknown $z_0$ with ${\hat{z}_0}$.

\vspace{3mm}

\textbf{Gradient Guidance} leverages various model's information $p(y|x_t)$ to guide the generation process. Prior works\cite{dhariwal2021diffusion, sauer2023adversarial} demonstrated that classifiers can serve as effective guidance tools within diffusion models. Furthermore, \textit{Universal Guidance for Diffusion}(UGD) \cite{bansal2023universal} extends this concept by employing the initial state $\hat{x}_0$ instead of $x_t$ from the pixel space as input to the guidance function $f$. This approach generalizes the guidance mechanism sohat a guidance function trained in the pixel space does not need to be rere-trained in the latent space. With this mechanism, UGD applies various guidance functions to diffusion such as segmentation, object detection, face recognition, etc.

\vspace{1.mm}
\subsection{Multi-modal Guidance for Diffusion} 

\textbf{Visual Guidance via Discriminator.} For visual guidance, we employ the discriminator architecture from StyleGAN-T \cite{sauer2023stylegan}, trained on a custom dataset we generated, with mean square error (MSE) as the loss function $\ell$. Following the Forward Guidance outlined in the Universal Guidance Diffusion \cite{bansal2023universal}, we apply the discriminator guidance during inference. We define $c$ as the text prompt, which provides semantic guidance during the generation process. The variable $y$ is set to a vector of ones, matching the output dimension of the discriminator $f_D$, to enforce real hand guidance. The guidance strength is modulated by the scalar weight $w$. Following the approach of SDEdit \cite{meng2021sdedit}, we apply guidance selectively at specific time steps (650 to 150) rather than at every iteration. The denoising function $\epsilon_{\theta_{V}}$ is formulated as follows:
\begin{equation}
\epsilon_{\theta_{V}}(z_t,c, t)=\epsilon_{\theta}(z_t,c,t) + w \cdot \nabla_{z_t} \ell(y, f_D(\hat{x}_{0},c))
\end{equation}

\vspace{3mm}

\textbf{Textual Guidance via LoRA.} We apply textual guidance to MGHanD using the method from ConceptSlider\cite{gandikota2023concept}. ConceptSlider is a diffusion model that enables detailed control over target attributes by applying a LoRA\cite{hu2021lora} Adapter to the frozen weights of Stable Diffusion. To enhance the hand representation, we used a neutral prompt $p$ as ‘hands’ and positive prompts $p_{+_i}$ such as ‘realistic', 'five fingers’ and negative prompts $p_{-_j}$ as 'distorted', 'malformed hands' during training. To learn the directional shift that brings $p$ closer to $p_+$ while pushing it away from $p_-$, we trained the model using $l_2$ loss. The intensity of hand correction was controlled through $v$.
\begin{equation}
\epsilon_{\theta}(z_t, p, t) \leftarrow \epsilon_{\theta}(z_t, p, t) + v \left[  \epsilon_{\theta}(z_t, p_{+}, t) - \epsilon_{\theta}(z_t, p_{-}, t) \right]
\end{equation}

 \noindent In the inference steps, for various hand action prompts $c$, LoRA weights that guide the direction towards $c_+$ were added to the Stable Diffusion weights by a factor of $v$.

\begin{equation}
\epsilon_{\theta_{T}}(z_t, c, t) = \epsilon_{\theta}(z_t, c, t) + v \left( \epsilon_{\theta}(z_t, c_{+}, t) \right)
\end{equation}

\begin{figure*}
    \centering
    \includegraphics[width=1\textwidth]{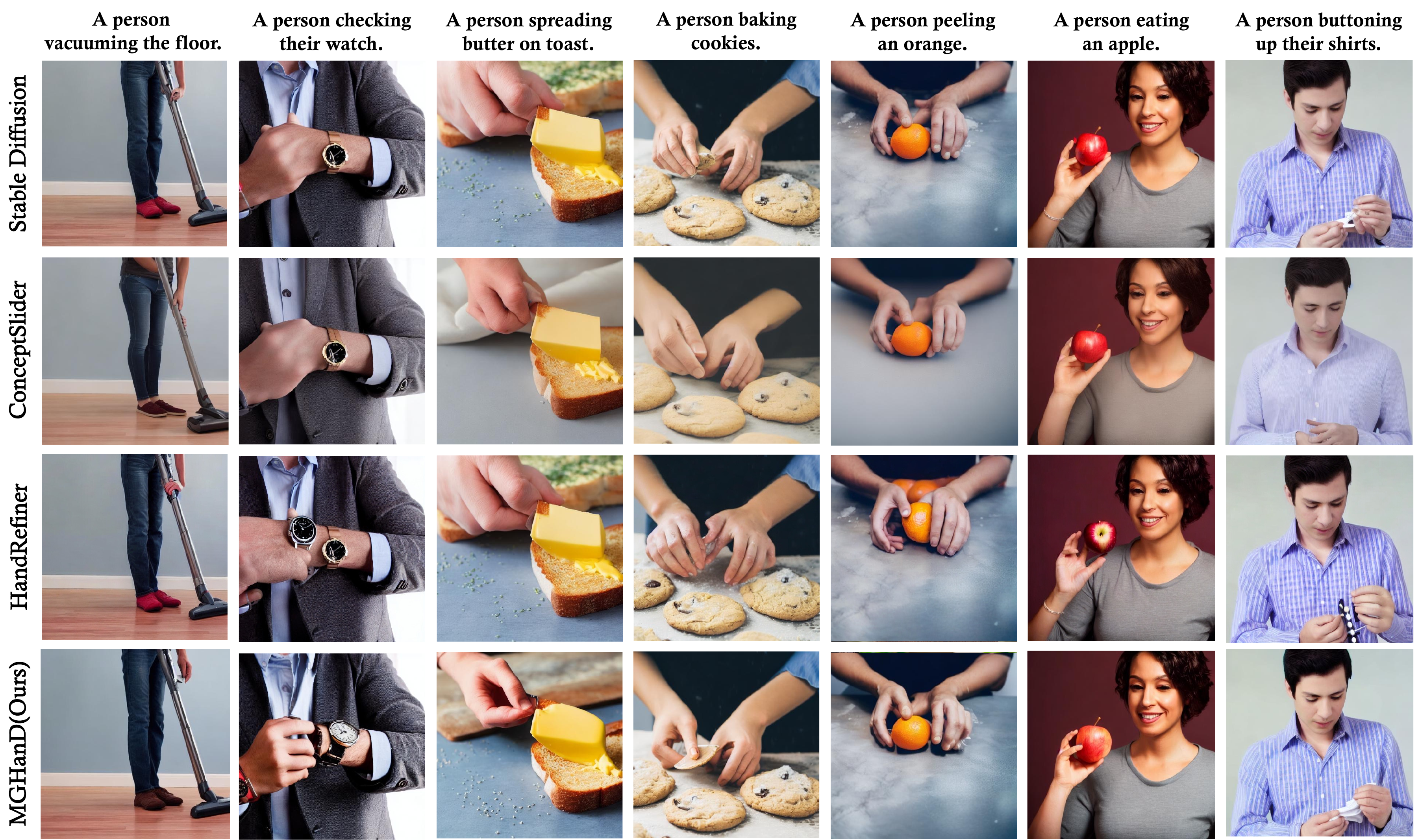} 
    \caption{\textbf{Qualitative Results: MGHanD vs. existing models.} Comparison between our MGHanD (bottom row) and existing methods (SD, ConceptSlider, HandRefiner) for six common hand-object interaction tasks.}
    \label{fig:qual_results}
\end{figure*}

 However, as illustrated in Fig.~\ref{fig:qual_results}, while the LoRA adaptor successfully guided the model towards improved hand generation, it introduced the unintended artifacts of background blurring at times. We incorporate the cumulative masking techniques outlined in the subsequent section to overcome this issue.

\renewcommand{\arraystretch}{1.4} 

\begin{table*}[ht]
\centering
\makebox[\textwidth]{
\begin{tabular}{l >{\centering\arraybackslash}p{2.cm} >{\centering\arraybackslash}p{2.2cm} >{\centering\arraybackslash}p{2.2cm} >{\centering\arraybackslash}p{2.2cm} >{\centering\arraybackslash}p{2.2cm}} 
\Xhline{2\arrayrulewidth}
\textbf{Method} & \textbf{FID\textcolor{blue}{$\downarrow$}} & \textbf{KID\textcolor{blue}{$\downarrow$}} & \textbf{Hand Conf.\textcolor{red}{$\uparrow$}} & \textbf{Hand Prob.\textcolor{red}{$\uparrow$}} & \textbf{CLIP Sim.\textcolor{red}{$\uparrow$}} \\ 
\Xhline{1\arrayrulewidth}
Stable Diffusion \cite{rombach2022high} & 1.0438 & 0.1476 & 0.8965 & 0.6708 & 29.8231 \\ 
ConceptSlider\cite{gandikota2023concept} & 4.5134 & 0.1445 & 0.8790 & \underline{0.7208} & \textbf{30.5508} \\ 
HandRefiner\cite{lu2023handrefiner} & \underline{0.9695} & \underline{0.1427} & \underline{0.8969} & 0.64 & \underline{30.35} \\ 
\Xhline{0.5\arrayrulewidth} 
MGHanD (Ours) & \textbf{0.9601} & \textbf{0.1368} & \textbf{0.9009} & \textbf{0.7250} & 30.1349 \\ 
\Xhline{0.5\arrayrulewidth} 
\Xhline{2\arrayrulewidth} 
\end{tabular}}
\caption{\textbf{Comparison of Methods in Perceptual Similarity Measures.} FID\cite{heusel2017gans} and KID\cite{binkowski2018demystifying} represent the diversity and fidelity of generated images, where lower values indicate better performance. Hand Confidence measures the quality of detected hands, while Hand Probability reflects the likelihood of hand detection in the images; both metrics are computed using Mediapipe\cite{lugaresi2019mediapipe}. CLIP\cite{radford2021learning} Similarity assesses the alignment between the prompt and the generated image, considering both the object and the background.}
\label{tab:comparison}
\end{table*}

\subsection{Cumulative Hand Mask}

To enhance the accuracy of hand region detection, we employ a dynamically evolving binary mask \( M_t \), which is updated at each time step based on the predictions the detection model \cite{lugaresi2019mediapipe}. The mask is defined as follows:

\begin{equation}
M_t \in \{0,1\}^{H \times W}
\end{equation}

\noindent where \( H \) and \( W \) denote the image dimensions. Each pixel \( M_{t}(i,j) \) is set to 1 if classified as part of the hand region and remains 0 otherwise. The mask expands progressively as newly detected hand regions are integrated.

To mitigate false positives, we apply a confidence threshold \( \tau=0.4 \), ensuring that only high-confidence detections contribute to mask expansion:

\begin{equation}
M_{t+1} = \max(M_t, \mathbb{I}(S_t \geq \tau) \cdot D_t)
\end{equation}

\noindent where \( S_t \) represents the detection confidence, and \( \mathbb{I}(\cdot) \) is an indicator function.  
\( D_t \) denotes the binary hand detection result at timestep \( t \), where \( D_t(i,j) = 1 \) if the detection model identifies a hand at pixel \( (i, j) \), and \( 0 \) otherwise.  
This cumulative mask \( M_t \) ensures that guidance remains focused on the hand region while minimizing background interference.





\begin{figure}[H]
        \centering
        \includegraphics[width=1\linewidth]{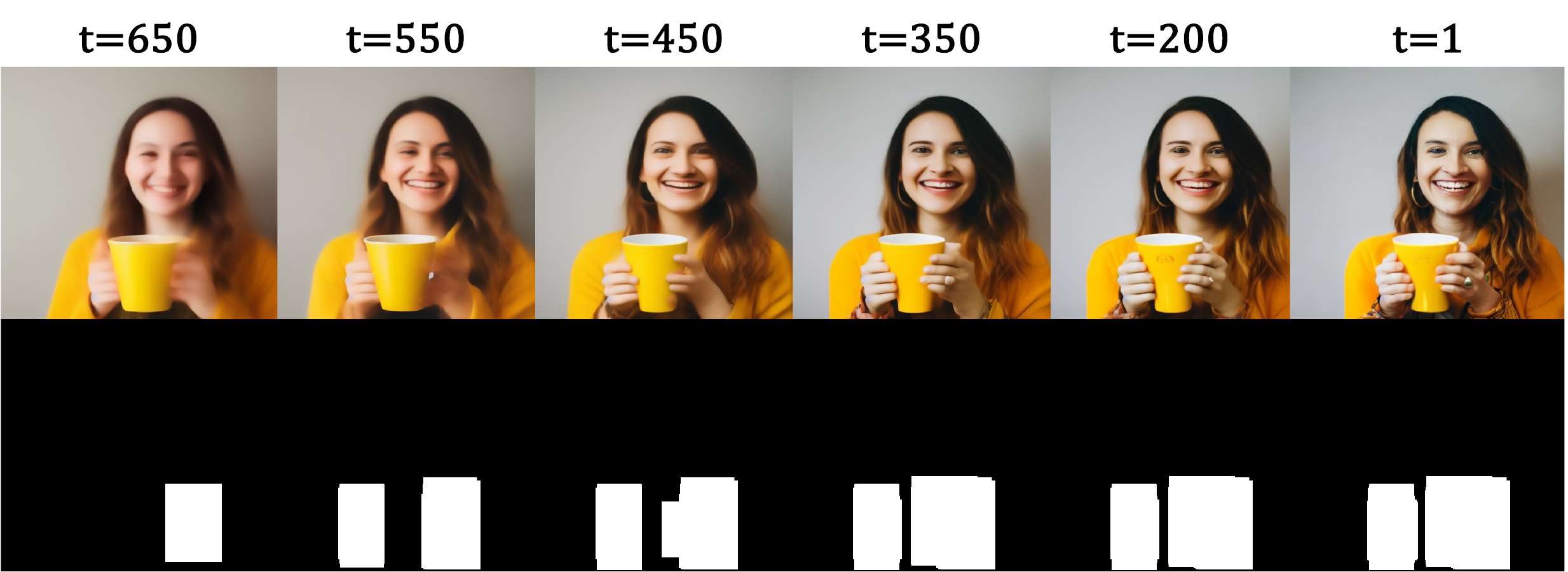}
        \caption{Visualization of the diffusion process's effect on the cumulative mask. By accumulating the mask over time steps, the method is robust to occasional errors in hand detection.}
    \label{fig:data_generation}
\end{figure}

\vspace{1.5mm}

\section{Experiments}

This section evaluates our MGHanD model. We describe the datasets (\ref{subsec:datasets}), implementation (\ref{subsec:implementation}), evaluation metrics (\ref{subsec:metrics}), and results (\ref{subsec:results}). Our experiments show MGHanD's ability to generate quality images with realistic hands from text prompts.

\vspace{1.5mm}
\subsection{Datasets}\label{subsec:datasets}

In this study, we developed two guidance models with different modalities using a custom dataset. For the Hand Discriminator model, we created a dataset comprising real hand images, generated hand images, and corresponding captions. The real hand images were sourced from two primary datasets: 100DOH\cite{Shan20}, which includes 27.3K videos and 100K frames featuring hands and hand-object interactions in both first-person and third-person perspectives, and HAGRID\cite{kapitanov2024hagrid}, a third-person dataset containing 554K images across 18 hand gesture classes. To ensure data quality, we filtered these datasets using a Mediapipe\cite{lugaresi2019mediapipe} hand detection score threshold of 0.8, excluding low-quality images. Captions were generated using the LLaVa\cite{llava2023} model, and Stable Diffusion was used to create generated images for fake data. This process resulted in 1,548 images from 100DOH and 1,626 from HAGRID to ensure that third-person images did not dominate. For textual guidance, we followed the ConceptSlider\cite{gandikota2023concept} framework, using 'hands' as the neutral prompt and creating five prompt sets by combining positive terms such as 'realistic hands', 'five fingers', '8K', and 'correct anatomy'. Negative terms included 'distorted hands', 'clumsy hands', 'poorly drawn hands', and 'blurry hands'.

\vspace{1.5mm}

\subsection{Implementation Details}\label{subsec:implementation}

We trained the Hand Discriminator to generate effective guidance for both images and prompts. For text encoding, we utilized the CLIP-ViT-L/16 model\cite{radford2021learning}, while for image encoding, we employed the DINOv2-ViT-S/14\cite{oquab2023dinov2} model. Additionally, we applied data augmentation \cite{zhang2023adding} before the feature extraction network in the discriminator. The model was trained for 120 epochs on a single A100 GPU with a batch size of 64. The Textual Guidance LoRA adapter was applied to the Stable Diffusion v1-4 checkpoint and trained for 1000 epochs with a batch size of 1, using the AdamW optimizer with a learning rate of \(10^{-4}\). The LoRA rank was set to 4.

\textbf{MGHanD Inference.} We apply multi-modal guidance to the Stable Diffusion v1-4 checkpoint to generate 512x512 size images. The DDIM \cite{song2020denoising} sampler is used with 100 steps, and the classifier-free guidance (cfg) weight is set to a default value of 3. To maintain the semantics and structure of the original pre-trained model for efficiency, the two guidance mechanisms are not applied from the first time step T. Instead, the Discriminator and LoRA Adapter is scaled and applied starting from step 65. 

\vspace{1.mm}
\subsection{Evaluation Metrics}\label{subsec:metrics}
We evaluate the quality of images generated by MGHanD using several metrics. To assess the overall image quality and diversity, we use the Frechet Inception Distance (FID)\cite{heusel2017gans} and Kernel Inception Distance (KID) \cite{binkowski2018demystifying}, computed against a reference set of  HAGRID\cite{kapitanov2024hagrid} test dataset. For hand-specific quality, we measure Hand Confidence obtained from the hand detector \cite{lugaresi2019mediapipe}. Hand Probability measures the likelihood that a hand is properly generated in images where the prompt explicitly requires its presence. Finally, we use the CLIP Similarity\cite{radford2021learning} to evaluate text-image consistency, ensuring that the generated images accurately reflect the provided textual descriptions. 

\vspace{1.mm}
\subsection{Results}\label{subsec:results}
\subsubsection{Quantitative Results and Ablation Studies}
Comparison of Methods and Other Models
\textbf{Quantitative Results} To evaluate the effectiveness of our proposed MGHanD model, we conducted a comprehensive comparison against three established methods: Stable Diffusion, HandRefiner, and ConceptSlider. The Stable Diffusion model, trained on the general-purpose LAION-5B dataset\cite{schuhmann2022laion}. HandRefiner is inpainting model based on ControlNet that offers some improvement by depth image conditioning at the hand-detected area, while ConceptSlider provides another approach with latent diffusion steering to better direction with LoRA Adapter. The results are summarized in Table~\ref{tab:comparison}.
MGHanD achieves the lowest FID and KID scores, indicating superior overall image quality compared to other state-of-the-art models. For hand-specific quality, MGHanD also achieves the highest Hand Confidence score of 0.901\% and the highest Hand Probability of 0.725\%, which are 8\% and 13\% higher than those of Stable Diffusion and HandRefiner, respectively. This demonstrates MGHanD’s ability to consistently generate more realistic and accurate hand images. Additionally, the CLIP Similarity score for MGHanD is improved compared to baseline Stable Diffusion, indicating better text-prompt alignment.

\vspace{1.mm}

\textbf{Qualitative Results} Fig.~\ref{fig:qual_results} demonstrates the comparative performance of hand generation models. While ConceptSlider and HandRefiner improve upon Stable Diffusion's hand representations, they exhibit notable limitations. ConceptSlider often produces unnatural hand shapes, as seen in the \textit{``checking their watch"} image, while HandRefiner occasionally fails to detect hands, leaving them unmodified, as in the \textit{``spreading butter"} image. 

In contrast, MGHanD effectively detects and refines even small hand regions, as demonstrated in the first-column examples. Additionally, it successfully corrects images where fingers are not clearly separated, as seen in the second and third examples. Moreover, MGHanD accurately reconstructs anatomically plausible thumbs in cases where unnatural thumb structures are generated, as illustrated in the fourth and fifth examples. Furthermore, as shown in the sixth and seventh examples, our model ensures natural hand corrections not only in first-person perspectives but also in third-person viewpoints.

\begin{figure}
    \centering
    \includegraphics[width=1.0\linewidth]{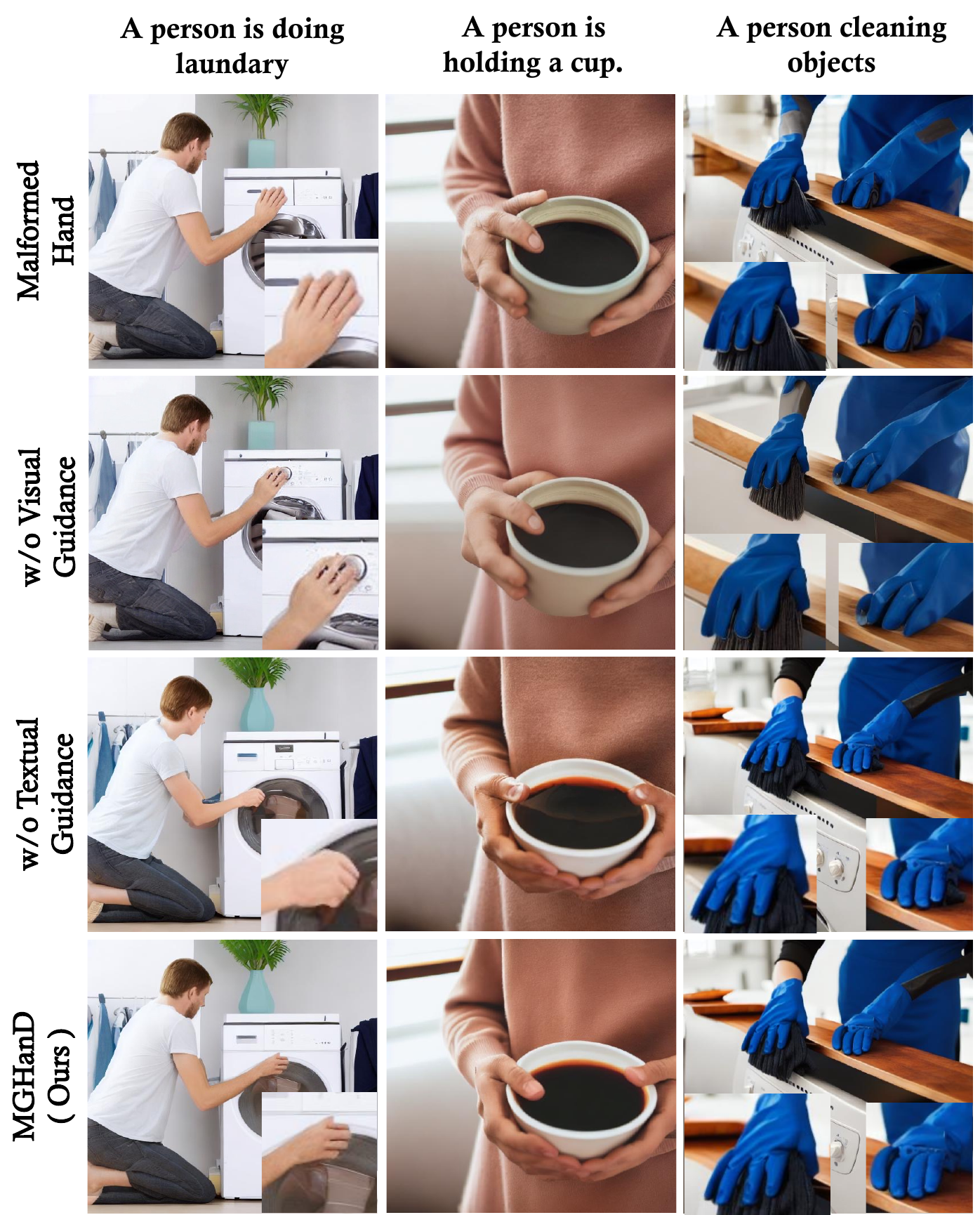}
    \caption{\textbf{Ablation study on Multi-modal Guidance.} Comparison of original images, MGHanD model, and versions without visual or textual guidance, highlighting the impact of each component.}
    \label{fig:ablation_guidance}
\end{figure}

\begin{figure}
    \centering
    \includegraphics[width=1.0\linewidth]{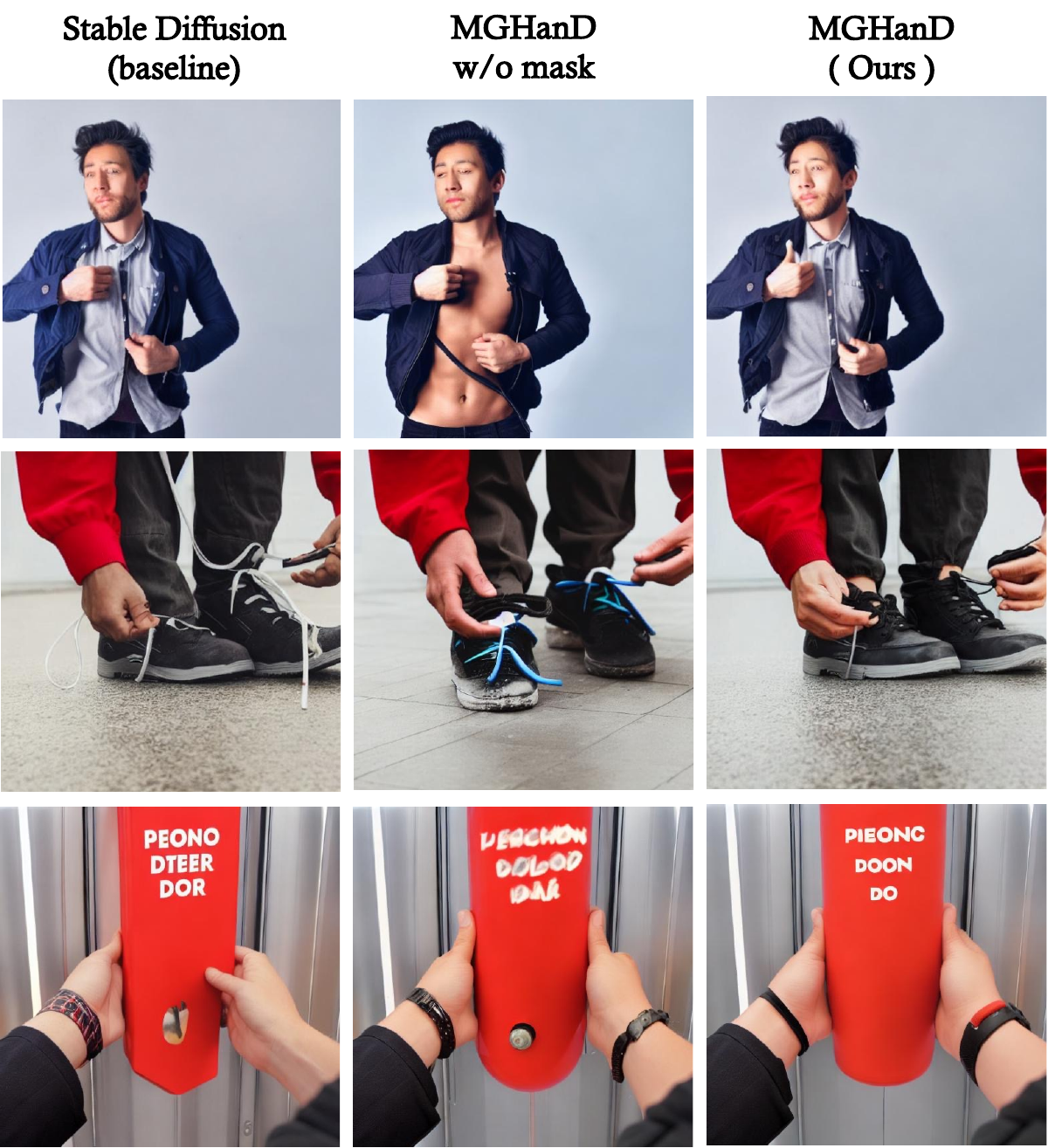}
    \caption{\textbf{Ablation with Cumulative Hand Mask.} Comparison of original images, MGHand without cumulative hand mask, and MGHanD with mask.}
    \label{fig:ablation_mask}
\end{figure}
\vspace{1.mm}


\textbf{Ablation Studies} Since MGHanD aims to refine intricate hand details, we primarily focus on qualitative ablation studies rather than quantitative metrics. As shown in Figure~\ref{fig:ablation_guidance}, the impact of each guidance mechanism is evident. Without Visual Guidance, the overall image remains largely unchanged, but the hand structure often appears unnatural, as seen in the \textit{“doing laundry”} and \textit{“holding a cup”} examples. In contrast, the absence of Textual Guidance leads to significant artifacts, such as missing fingers or awkward hand-object interactions. For instance, in the \textit{“doing laundry”} example, the thumb disappears, and in the \textit{“cleaning objects”} example, the gloves appear distorted, failing to align naturally with the hands. These qualitative findings highlight the necessity of both visual and textual guidance to ensure anatomically accurate and contextually coherent hand generation.

We also conducted an ablation study on the Cumulative Hand Mask. As depicted in \ref{fig:ablation_mask}, when the mask was not applied, unintended modifications occurred, such as the disappearance of objects, color shifts, and loss of fine details along with hand corrections. By incorporating the Cumulative Hand Mask, we ensured that while the hands were refined, the surrounding objects and background retained their realistic details.

\begin{figure}[H]
        \centering
        \includegraphics[width=1.0\linewidth]{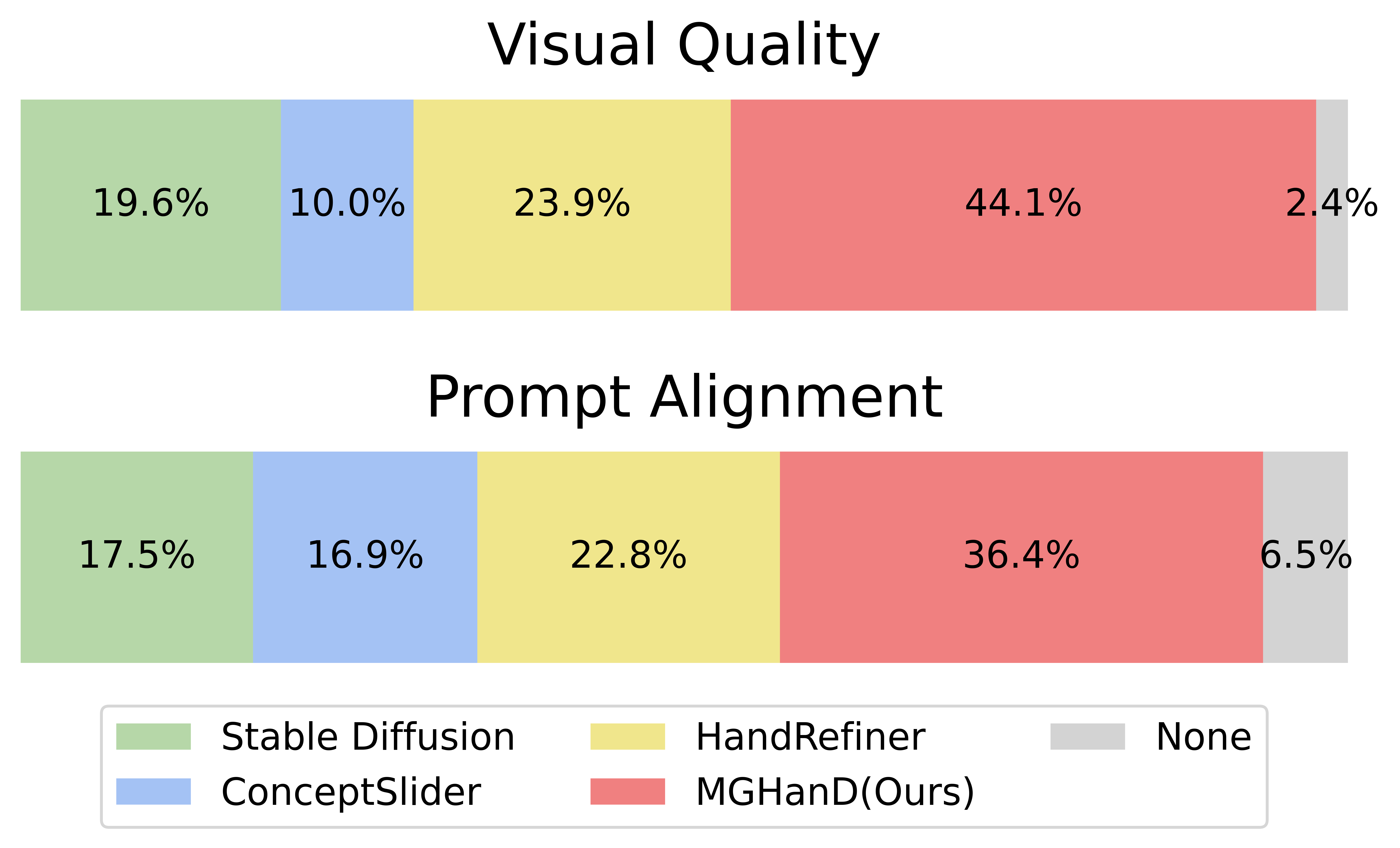}
        \caption{\textbf{Human evaluation} results comparing the visual quality and prompt alignment of images generated by different methods. MGHanD (Ours) outperforms other approaches in both metrics, demonstrating its superiority in generating visually appealing and semantically relevant images.}
    \label{fig:user_study}
\end{figure}

\vspace{1.mm}
\textbf{User Study}
We conducted a comparative user study to evaluate four image generation models: Stable Diffusion, ConceptSlider, HandRefiner, and our proposed model, MGHanD, with an additional `None' option if no preference was discernible. The study involved 40 participants assessing images generated from 20 uniformly distributed prompts across all models. Participants evaluated pairs of images based on two criteria: visual quality, focusing on the accurate representation of hands, and prompt alignment, assessing how well the generated images matched the given text descriptions. As shown in Fig.~\ref{fig:user_study}, MGHanD demonstrated superior performance in both categories. For visual quality, MGHanD was preferred in 44.1\% of comparisons, outperforming other models (HandRefiner: 23.9\%, Stable Diffusion: 19.6\%, ConceptSlider: 10.0\%, None: 2.4\%). In prompt alignment, MGHanD again led with 36.4\% of preferences, surpassing other models (HandRefiner: 22.8\%, Stable Diffusion: 17.5\%, ConceptSlider: 16.9\%, None: 6.5\%). These results highlight MGHanD's ability to generate images with higher visual appeal, particularly in hand representation, and better alignment with given prompts.

\section{Limitations and Future Work}

While our approach effectively enhances hand generation, there remain areas for further refinement. First, the current generation time of approximately 2 minutes per image is primarily influenced by the guidance process in latent space. While this ensures high-quality hand synthesis, optimizing inference efficiency could improve practicality. Second, minor artifacts may arise in certain cases, such as unintended object alterations when expanding mask ranges or occasional challenges in precisely controlling the number of hands. These occurrences highlight the complexity of maintaining both structural consistency and contextual precision. To address these aspects, we propose (a) refining masking techniques for more precise guidance control and (b) optimizing model efficiency to reduce generation time while preserving image quality. These improvements could contribute to a more robust and efficient framework, aligning generated images more closely with user expectations.

\section{Conclusion}

We have developed an end-to-end model generating high-quality hands from text prompts by simultaneously applying novel Visual and Textual Guidance to a diffusion model. To train the Visual Guidance model, we created a compact dataset consisting of 6.3K real and fake captioned images. Additionally, a Cumulative Hand Mask was applied to preserve the broad generative capabilities of Stable Diffusion while refining the hands, utilizing a cumulative approach across time steps for more stable masking in the generated images. As a result, we achieved quantitative and qualitative results and further demonstrated superior performance compared to other models in a user study.

\vspace{2mm}




\end{document}